\documentclass[10pt,twocolumn,letterpaper]{article}

\usepackage[pagenumbers]{iccv} 

\newcommand{\red}[1]{{\color{red}#1}}

\definecolor{iccvblue}{rgb}{0.21,0.49,0.74}
\usepackage[pagebackref,breaklinks,colorlinks,allcolors=iccvblue]{hyperref}
\usepackage{multirow} 
\newcommand{\OurMethod}[0]{\textit{{ColorFlow}}}
\newcommand{\BenchmarkName}[0]{\textit{{ColorFlow-Bench}}}
\def\paperID{****} 
\def\confName{****}
\def\confYear{2025}

\title{ColorFlow:  Retrieval-Augmented Image Sequence Colorization }

\author{
Junhao Zhuang$^{1, 2*}$ \quad Xuan Ju$^{2*}$ \quad Zhaoyang Zhang$^{2\dag}$ \quad  Yong Liu$^{1, 2}$ \quad Shiyi Zhang$^{1, 2}$\\
Chun Yuan$^{1\ddag} $ \quad Ying Shan$^{2\ddag} $\\
{
$^{1}$Tsinghua University \ 
$^{2}$ARC Lab, Tencent PCG \
}
}

\begin{document}
\twocolumn[{%
\renewcommand\twocolumn[1][]{#1}%
\maketitle
\begin{center}
    \vspace{-8mm}
    \centering
    \includegraphics[width=0.98\textwidth]{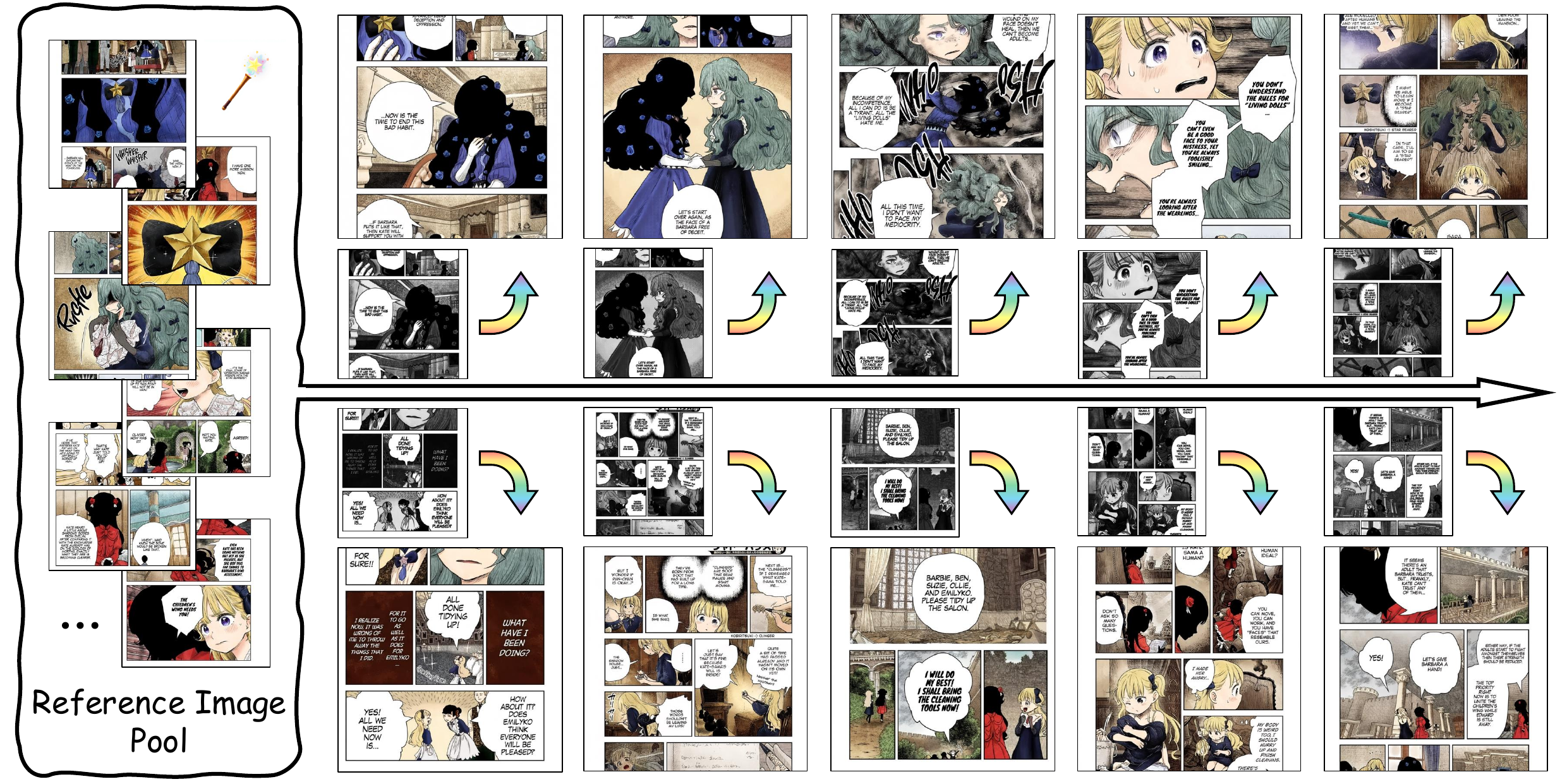}
    \vspace{-2mm}
    \captionof{figure}{
        \label{fig:teaser}
    \textbf{ColorFlow} is the first model designed for fine-grained ID preservation in image sequence colorization, utilizing contextual information. Given a reference image pool, ColorFlow accurately generates colors for various elements in black and white image sequences, including the hair color and attire of characters, ensuring color consistency with the reference images. [Best viewed in color with zoom-in]. }
    \vspace{-1mm}
\end{center}%
}]
\let\thefootnote\relax\footnotetext{
$^*$ Equal Contribution. \quad $^\dag$ Project lead. \quad $^\ddag$ Corresponding authors.}

\begin{abstract}

\vspace{-3mm}

Automatic black-and-white image sequence colorization while preserving character and object identity (ID) is a complex task with significant market demand, such as in cartoon or comic series colorization. Despite advancements in visual colorization using large-scale generative models like diffusion models, challenges with controllability and identity consistency persist, making current solutions unsuitable for industrial application.
To address this, we propose ColorFlow, a three-stage diffusion-based framework tailored for image sequence colorization in industrial applications. Unlike existing methods that require per-ID finetuning or explicit ID embedding extraction, we propose a novel robust and generalizable Retrieval Augmented Colorization pipeline for colorizing images with relevant color references.
Our pipeline also features a dual-branch design: one branch for color identity extraction and the other for colorization, leveraging the strengths of diffusion models. We utilize the self-attention mechanism in diffusion models for strong in-context learning and color identity matching.
To evaluate our model, we introduce ColorFlow-Bench, a comprehensive benchmark for reference-based colorization. Results show that ColorFlow outperforms existing models across multiple metrics, setting a new standard in sequential image colorization and potentially benefiting the art industry. Code and model will be released.
\end{abstract}    
\vspace{-5mm}
\section{Introduction}
\label{sec:intro}
\vspace{-1mm}

Diffusion models have made substantial progress in generation, achieving state-of-the-art results in controllable image generation, including image inpainting\cite{zhuang2023task,ju2024brushnet}, image colorization\cite{zabari2023diffusing,weng2024cad}, and image editing~\cite{brooks2023instructpix2pix}. This progress has sparked the growth of numerous downstream tasks.
However, there has been limited attention to a problem where diffusion-based generation could significantly reduce labor costs: reference-based image sequence colorization, which can be used in manga creation, animation production, and black-and-white film colorization.

%
%

Recently, with the unprecedented image generation capabilities of diffusion models, there is growing interest in colorization using diffusion models\cite{cao2023animediffusion, zhang2023adding, zabari2023diffusing}. However, most efforts\cite{zhang2023adding,liang2024control,liu2023improved,zabari2023diffusing, weng2024cad} only consider basic text-to-image settings without reference to color, which is far from practical application.
While some research\cite{cao2023animediffusion, liu2025manganinja, li2022eliminating} explores reference-based colorization, three key challenges remain: \textbf{1)} selecting suitable reference image sets is complex and time-consuming, \textbf{2)} models are limited to single-image references, restricting multi-object representation, and \textbf{3)} latent-diffusion\cite{ldm22} frameworks often produce structural distortions due to VAE compression, affecting texture and text details.

To tackle these challenges, we introduce \OurMethod~, a novel three-stage framework for multi-reference image sequence colorization that preserves multiple color IDs. It consists of:
\textbf{a)}\textbf{ Retrieval-Augmented Pipeline (RAP)}: Automatically matches patches from a black-and-white image to a reference pool, providing accurate color IDs with enhanced usability and eliminating the need for labor-intensive reference image selection.
\textbf{b)} \textbf{In-context Colorization Pipeline (ICP)}: Employs a convolutional branch and a LoRA-fine-tuned diffusion model to integrate reference color information, leveraging contextual learning for precise multi-ID matching without requiring supervised ID correspondence training.
\textbf{c)} \textbf{Guided Super-Resolution Pipeline (GSRP)}: Minimizes output distortions by fusing high-quality black-and-white features with low-quality colored image features in the VAE encoder, utilizing residual connections to achieve high-fidelity color outputs.

To ensure a comprehensive evaluation of \OurMethod~, we construct \BenchmarkName, a benchmark comprising 30 manga chapters, each containing 50 black-and-white images and 40 reference images. Experimental results indicate that \OurMethod~ achieves state-of-the-art performance across five metrics in both pixel-wise and image-wise evaluations. Compared to existing works, \OurMethod~ excels in preserving finer-grained color identities within image sequences, leading to significant enhancements in image quality. Our contributions are summarized as follows:
\begin{itemize}[itemsep=-0.1em] 
    \item We introduce \OurMethod~, a novel three-stage framework that effectively facilitates fine-grained color ID from retrieved multi-references through contextual learning, enabling high-quality colorization of image sequences.
    
    \item We establish \BenchmarkName~, a comprehensive benchmark specifically designed for multi-reference-based image sequence colorization.
    
    \item Extensive evaluations demonstrate that our method surpasses existing approaches in both perceptual metrics and subjective user studies. Our model achieves over 37\% reduction in FID metrics compared to state-of-the-art colorization models. Additionally, our proposed model ranks first in user study scores for aesthetic quality, similarity to the reference, and sequential consistency.
\end{itemize} 
\vspace{3mm}
\section{Related Work}
\label{sec:related_work}

\textbf{Image Colorization}~\cite{zhang2018two,comicolorization} aims to transform grayscale images (\textit{e.g.}, manga~\cite{qu2006manga}, line art~\cite{kim2019tag2pix}, sketches~\cite{zhang2023adding}, and grayscale natural images~\cite{zabari2023diffusing}) into their colored counterparts. 
To enhance controllability, various conditions are used to imply color information, including scribbles~\cite{dou2021dual,zhang2021user,zhang2018two,yun2023icolorit,ci2018user,liu2018auto,sangkloy2017scribbler,zhang2017real,frans2017outline,sykora2009lazybrush,qu2006manga}, reference images~\cite{cao2023animediffusion,zou2024lightweight,wu2023flexicon,wang2023unsupervised,wu2023self,bai2022semantic,li2022eliminating,li2022style,zhang2022scsnet,yin2021yes,li2021globally,zhang2021line,lu2020gray2colornet,chen2020active,akita2020colorization,xie2020manga,xu2020stylization,lee2020reference,sun2019adversarial,li2019automatic,fang2019superpixel,xiao2020example,he2018deep,xian2018texturegan,comicolorization}, palettes~\cite{utintu2024sketchdeco,wu2023flexicon,wang2022palgan,xiao2020example,bahng2018coloring,chang2015palette}, and text~\cite{zabari2023diffusing,zhang2023adding,weng2024cad,chang2023coins,chang2022coder,weng2022code,cao2021line,zou2019language,kim2019tag2pix,manjunatha2018learning,bahng2018coloring,chen2018language}.
Specifically, scribbles provide simple and freehand color strokes as color pattern hints. The Two-stage Sketch Colorization~\cite{zhang2018two} employs a two-stage CNN-based framework that first applies strokes of color over the canvas, then corrects color inaccuracies and refines details. Reference image-based colorization transfers colors from a reference image that contains similar objects, scenes, or textures. ScreenVAE~\cite{xie2020manga} and Comicolorization~\cite{comicolorization} compress color information from the reference image into a latent space, then inject the latent representation into the base colorization network. Palette-based models~\cite{chang2015palette,utintu2024sketchdeco} use the palette as a stylistic guide to inspire the overall color theme of the image.
With the emergence of diffusion models~\cite{ho2020denoising,song2020denoising}, text has become one of the most significant forms of guidance for image generation, and is thus widely used in image colorization. Text guidance uses a text prompt describing the desired color theme, object colors, or overall mood. ControlNet~\cite{zhang2023adding} adds additional trainable modules to pre-trained text-to-image diffusion models~\cite{ldm22} and leverages the native text-to-image capabilities of diffusion models for colorization.

However, whether using palettes, text, or latent representations, these methods can only provide a rough color style and cannot guarantee the accurate color preservation of instances in black-and-white image. In contrast, ColorFlow achieves instance-level color preservation across frames in image sequences by introducing a retrieval-augmented pipeline and a context feature matching mechanism.


\vspace{10pt}
\noindent\textbf{Image-to-image translation} aims to  establish a mapping from a source domain to a target domain (e.g., sketch-to-image~\cite{zhao2024uni,zhang2023adding}, pose-to-image~\cite{ju2023humansd,liu2023hyperhuman}, image inpainting~\cite{zhuang2023task,ju2024brushnet}, and image editing~\cite{ju2024pnp,brooks2023instructpix2pix}). Recent advancements in diffusion models~\cite{dai2023emu,ho2020denoising,song2020denoising,rombach2022high} have made them dominant in this task. Approaches are mainly categorized into inference-based~\cite{hertz2022prompt,avrahami2022blended} and training-based paradigms~\cite{zhang2023adding,ju2024brushnet}.
Inference-based methods often use a dual-branch structure~\cite{ju2024pnp}, where a source branch preserves essential content and a target branch maps images with guidance. These branches interact through attention or latent feature integration, but often suffer from insufficient control.
Training-based methods~\cite{zhang2023adding,zhao2024uni,mou2024t2i} are popular for their high quality and precise control. Stable Diffusion~\cite{rombach2022high} adds depth control by directly concatenating control conditions with noisy input and fine-tuning the model end-to-end. ControlNet~\cite{zhang2023adding} uses a dual-branch design to add control conditions to a frozen pretrained diffusion model, enabling plug-and-play control while preserving high image generation quality.

Notably, none of these approaches specifically address identity preservation across frames in sequential image translation tasks, which limits their applicability in practical industrial scenarios involving sequential images. In contrast, ColorFlow is designed to tackle this limitation, providing robust instance identity preservation in image sequence colorization tasks across frames.

\vspace{10pt}
\noindent\textbf{ID-Preservation} is a trending topic in the field of image generation. Previous approaches can be classified into two primary categories: the first involves fine-tuning generative models to enable them to memorize one or more predefined concepts~\cite{ruiz2023dreambooth,kumari2023multi,gal2022image}; the second employs plug-and-play modules that have been trained on large-scale datasets, allowing the model to control the generation of desired concepts using given image content during the inference stage~\cite{ye2023ip,li2024photomaker,wang2024instantid}. Generally, prior methods focus on a limited set of predefined concepts.

In contrast, we propose ColorFlow, which provides a robust and automated three-stage framework for sequential image colorization. 
ColorFlow effectively addresses the challenges of handling the dynamic and diverse characters, objects, and backgrounds present in comic sequences, making it well-suited for industrial applications.

\begin{figure*}[t]
    \centering
    \vspace{-3mm}
    \includegraphics[width=0.98\textwidth]{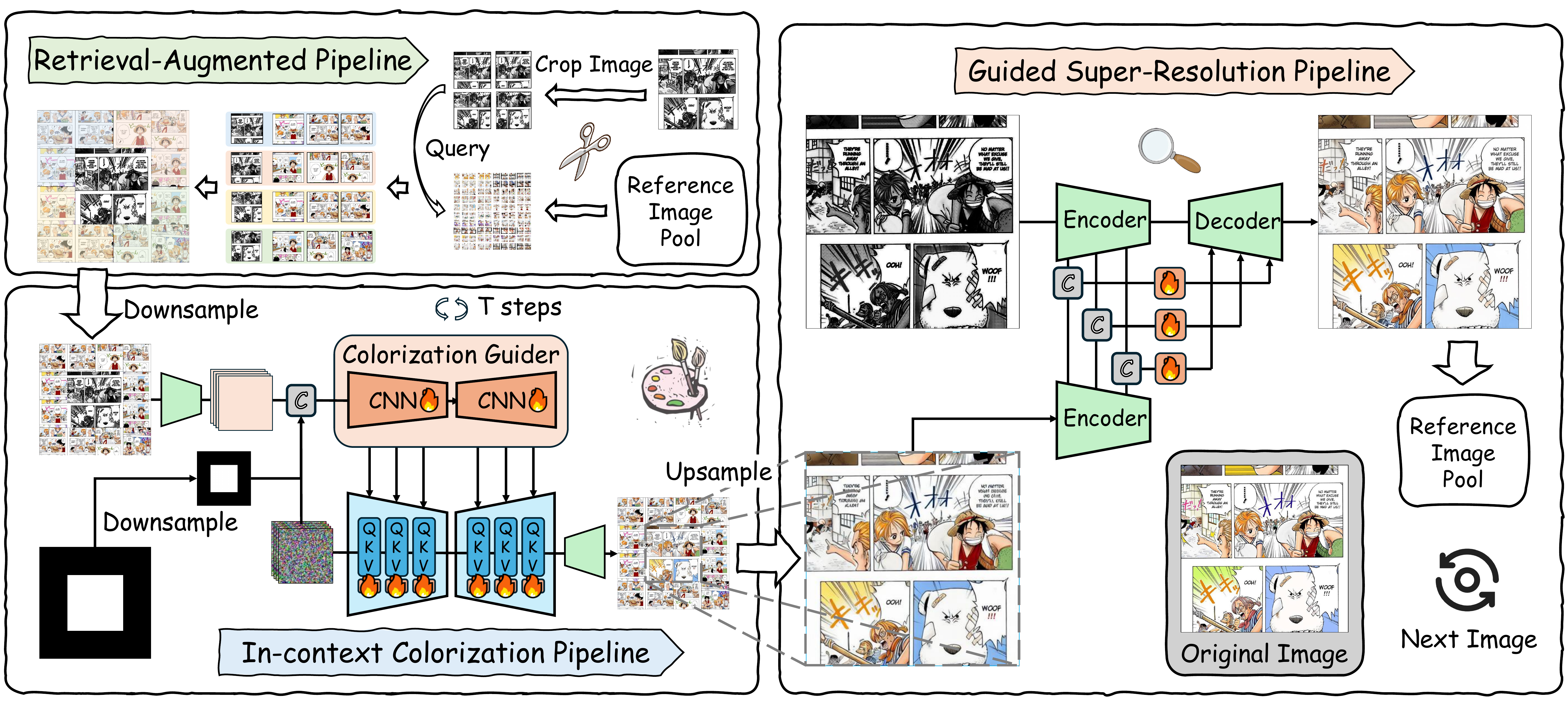}
    \vspace{-2mm}
    \caption{
    \textbf{The overview of ColorFlow}. This figure presents the three primary components of our framework: the Retrieval-Augmented Pipeline (RAP), the In-context Colorization Pipeline (ICP), and the Guided Super-Resolution Pipeline (GSRP). Each component is essential for maintaining the color identity of instances across black-and-white image sequences while ensuring high-quality colorization.
    }
    \vspace{-2mm}
    \label{fig:FlowChat}
\end{figure*}
\section{Method}
\label{sec:method}
Our aim is to colorize black-and-white images using colored images as references, ensuring consistency in characters, objects, and backgrounds throughout the image sequence. As illustrated in Figure~\ref{fig:FlowChat}, our framework consists of three main components: the Retrieval-Augmented Pipeline, the In-context Colorization Pipeline, and the  Guided Super-Resolution Pipeline. 





\subsection{Retrieval-Augmented Pipeline}
\label{sec:retrieval}
The Retrieval-Augmented Pipeline (RAP) is designed to identify and extract relevant colored references to guide the colorization process. 
To accomplish this, we first divide the input black-and-white image into four overlapping patches: top-left, top-right, bottom-left, and bottom-right. Each patch covers three-quarters of the original image's dimensions to ensure that important details are retained.
For each colored reference image, we create five patches: the same four overlapping patches and the complete image, providing a comprehensive set of reference data.

Next, we employ a pre-trained CLIP image encoder to generate image embeddings \( E_{bw} \) for the patches of the input image and \( E_{ref} \) for the reference patches. These embeddings are defined as follows:
\begin{equation}
E_{bw} = f_{CLIP}(P_{bw}) \quad \text{and} \quad E_{ref} = f_{CLIP}(P_{ref}),
\end{equation}
where \( P_{bw} \) represents the black-and-white patch and \( P_{ref} \) denotes the colored reference patch.

For each of the four patches from the input image, we compute the cosine similarity \( S \) between its embedding and the embeddings of the reference patches:
\begin{equation}
S(a, b) = \frac{a \cdot b}{\|a\| \cdot\|b\|}.
\end{equation}
where \( a \) and \( b \) are the embeddings of the query and reference patches, respectively. We define the top three similar patches for each query patch as follows:
\begin{equation}
\begin{aligned}
{Top}_3(E_{bw}^{(i)}) &= \{E_{ref}^{(j_1)}, E_{ref}^{(j_2)}, E_{ref}^{(j_3)} \mid \\
& j_k \in \arg\max_{k} S(E_{bw}^{(i)}, E_{ref}^{(k)}), \, k = 1, 2, 3 \},
\end{aligned}
\end{equation}
for \( i \in \{0, 1, 2, 3\} \), where \( E_{bw}^{(i)} \) denotes the embedding of the \( i \)-th query patch and \( E_{ref}^{(k)} \) denotes the embeddings of the corresponding reference patches.

After identifying the top three similar patches for each query region, we combine these selected patches into a unified output image. The patches corresponding to the top-left, top-right, bottom-left, and bottom-right regions are stitched together to create the composite image \( \mathcal{C}_{bw} \), as illustrated in Figure \ref{fig:FlowChat}. This spatial arrangement ensures the accurate placement of retrieved color information, enhancing the contextual relevance of the colorization process. 
In addition, we construct (  $\mathcal{C}_{color}$ ) by similarly stitching together the original colored versions corresponding to the black-and-white image patches. This forms a data pair with (  $\mathcal{C}_{bw}$ ) for subsequent colorization training.
By effectively gathering the most contextually relevant color information, the Retrieval-Augmented Pipeline sets the stage for the next stages of our framework, ensuring that the generated colors are harmonious and consistent with reference images.

\subsection{In-context Colorization Pipeline}
\label{sec:Colorization}
The In-context Colorization Pipeline is a fundamental component of our framework, designed to convert black-and-white images into full-color versions by utilizing contextual information from retrieved patches.
We introduce an auxiliary branch called the Colorization Guider, which aids in incorporating conditional information into the model. This branch is initialized by replicating the weights of all convolutional layers from the U-Net of the diffusion model.

The inputs to the Colorization Guider consist of the noise latent variable \( Z_t \), the output of the variational autoencoder \( \text{VAE}(\mathcal{C}_{bw}) \) for the composite image \( \mathcal{C}_{bw} \), and the downsampled mask \( M \). These components are concatenated to form a comprehensive input for the model. Features from the Colorization Guider are integrated progressively into the U-Net of the diffusion model, enabling a dense, pixel-wise conditional embedding. Furthermore, we utilize a lightweight LoRA (Low-Rank Adaptation) approach to fine-tune the diffusion model for the colorization task.
The loss function can be formalized as follows:
\begin{equation}
\mathcal{L}_{Color} = E_{t, \mathcal{C}_{bw}, \epsilon_t} \| \epsilon_t - \epsilon_\theta(\{ \text{VAE}(\mathcal{C}_{bw}), M, Z_t \}, t) \|^2_2.
\end{equation}
During training, \( Z_t \) is derived from \( \text{VAE}(\mathcal{C}_{color}) \) through the forward diffusion process.

This training objective allows the model to efficiently denoise the input latent space, gradually reconstructing the desired colored outputs from black-and-white inputs while being guided by reference images. 
Although we do not explicitly map instances from the colored reference images to those in the black-and-white images, the retrieval mechanism ensures that the reference images contain similar content. As a result, the model naturally learns to leverage contextual information from the retrieved references to accurately colorize the black-and-white images.

\paragraph{Timestep shifted sampling.}
Given that the colorization process is primarily determined during the higher timesteps, an emphasis on higher timesteps is important for generation. We modify our sampling strategy by adjusting timestep \( t' \):
\begin{equation}
t' = \frac{e^\mu}{e^\mu + \left( \frac{T}{t} - 1 \right)}T, \quad t \sim \mathcal{U}(0, T].
\end{equation}
In this work, we set $\mu$ to 1.5. This adjustment enables the model to emphasize these higher timesteps, thereby enhancing the effectiveness of the colorization process.
\begin{figure*}[hbtp]
    \centering
    \vspace{-2mm}
    \includegraphics[width=0.75\textwidth]{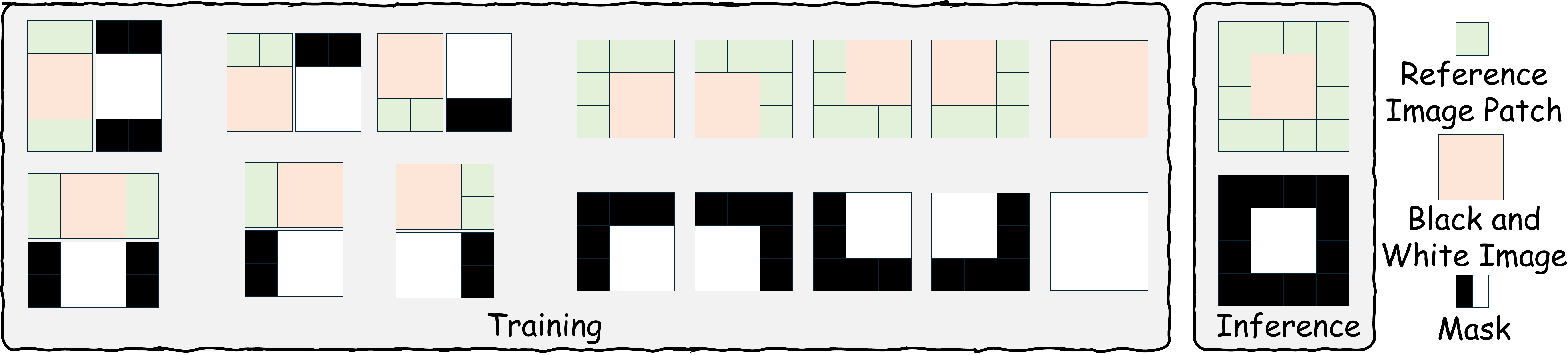}
    \caption{
    \textbf{Patch-Wise training strategy} is designed to reduce the computational demands of training on high-resolution stitched images. The left box displays segmented stitched images from the training phase, with the corresponding masks also segmented accordingly. The right box presents the complete stitched image and masks for the inference phase.}
    \label{fig:patch}
\end{figure*}
\begin{figure}[hbtp]
    \centering
    \includegraphics[width=0.45\textwidth]{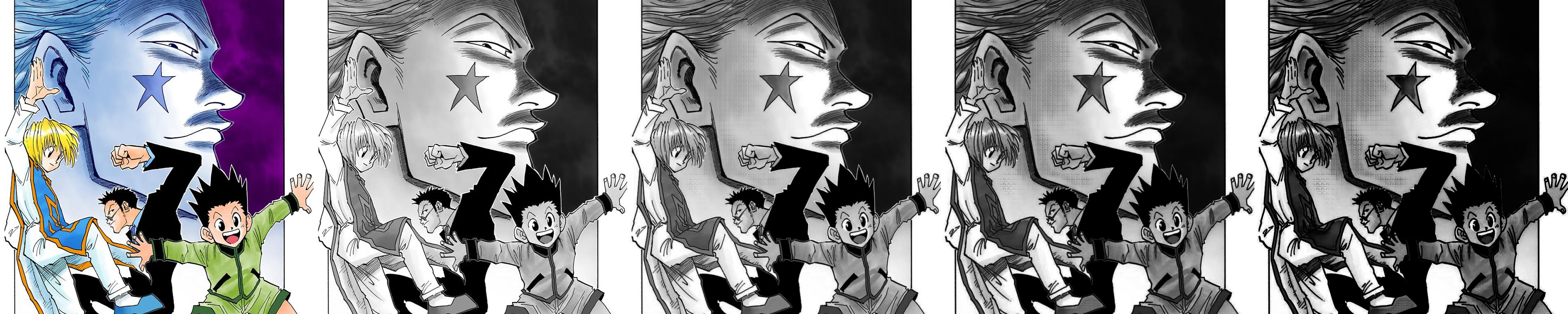}
    \caption{
    \textbf{Screenstyle augmentation}. From left to right: the colored manga, the grayscale manga, linear interpolations between the grayscale manga and the ScreenVAE~\cite{xie2020manga} output with proportions of 0.66 and 0.33, the ScreenVAE output.
    }
    \vspace{-0.4 cm}
    \label{fig:Mangastyle}
\end{figure}
\paragraph{Screenstyle augmentation.}
Xie et al. previously introduced ScreenVAE \cite{xie2020manga}, which enables the automatic conversion of colored manga into Japanese black-and-white styles. In this work, we augment the input images by performing random linear interpolation between the grayscale images and the outputs generated by ScreenVAE. This augmentation strategy, illustrated in Fig.~\ref{fig:Mangastyle}, helps the model better adapt to various styles and improves the overall performance of the colorization process.

\paragraph{Patch-Wise training strategy.}
To address the substantial computational demands of training on high-resolution stitched images, we introduce a patch-wise training strategy.
During training, we randomly crop segments from reference image patches, ensuring that the entire black-and-white image area is always included. The corresponding masks, which indicate the coloring regions, are cropped in the same manner. 
To further enhance performance, we downsample the input images, reducing computational load while preserving crucial details. This strategy significantly shortens training time per iteration, promoting faster model convergence. During inference, we use the complete stitched image to maximize the availability of contextual information for colorization, as shown in Fig.~\ref{fig:patch}.

\subsection{ Guided Super-Resolution Pipeline}
\label{sec:Super-Resolution}

The Guided Super-Resolution Pipeline is designed to tackle the challenges associated with downsampling during the colorization and to reduce the structural distortions often seen in the output from the latent decoder \( D \). These issues can significantly affect the quality of generated images.
This pipeline takes as input a high-resolution black-and-white image \( I_{bw}^{high} \) and a low-resolution colored output \( I_{color}^{low} \) produced by the In-context Colorization Pipeline. The goal is to produce a high-resolution colored image \( I_{pred}^{high} \). To achieve this, we first upsample the low-resolution colored image \( I_{color}^{low} \) to match the resolution of \( I_{bw}^{high} \) using linear interpolation. The upsampled colored image and the original high-resolution black-and-white image are then processed through the VAE encoder \( E \).

To enable effective feature integration, skip guidance is established between the encoder and decoder of the VAE. Intermediate features from both encoders are concatenated and passed to a fusion module \( F \), which transmits the combined information to the corresponding layers in the decoder. This multi-scale approach enhances detail restoration, as illustrated in Fig.~\ref{fig:FlowChat}.

The overall loss function for this process is defined as:
\begin{equation}
\begin{aligned}
\mathcal{L}_{SR} = & \mathbb{E}[| I_{bw}^{high} - D(F(\text{concat}(E_{features}(I_{bw}^{high}),  \\
& E_{features}(Upsample(I_{color}^{low})))), E(I_{bw}^{high}))|_1],
\end{aligned}
\end{equation}
where \( E_{features} \) denotes the intermediate features extracted from the VAE encoder. This pipeline effectively addresses the issues related to downsampling and structural distortions, resulting in a higher quality final output.

\vspace{-1mm}
\section{Experiments}
\label{sec:experiments}
\vspace{-1mm}

\subsection{Dataset and Benchmark}
\label{sec:dataset_and_benchmark}

\vspace{-1mm}

\begin{table*}[htbp]
    \centering
  \caption{
  \textbf{Quantitative comparisons with state-of-the-art models for Reference Image-based Colorization.} We compare two models without reference image input Manga Colorization V2 (MC-v2)~\cite{mcv2} and AnimeColorDeOldify (ACDO)~\cite{acdo}, and two reference image-based colorization models, Example Based Manga Colorization (EBMC)~\cite{isola2017image} and ScreenVAE~\cite{xie2020manga}. Best results are in \textbf{bold}.
  }
  \vspace{-2mm}
  \label{tab:exp1}
\resizebox{1.0\textwidth}{!}{
      \begin{tabular}{c|c|c c c c c|c c c c c}
    \toprule[1.5pt]
  \multirow{2}{*}{Method}  & \multirow{2}{*}{Reference-based} & \multicolumn{5}{c|}{Screenstyle}   & \multicolumn{5}{c}{Grayscale Image}\\
     &   & CLIP-IS$\uparrow$ & FID$\downarrow$  & PSNR$\uparrow$ &SSIM$\uparrow$ &AS$\uparrow$ &  CLIP-IS$\uparrow$ & FID$\downarrow$  & PSNR$\uparrow$ &SSIM$\uparrow$ &AS$\uparrow$\\ \midrule
    MC-v2 \cite{MangaColorV2} & &  0.8632 & 48.37 & 13.50 &0.6987 & 4.753 &  0.8833 & 33.14 & 17.20 &0.8396 & 4.845\\ \midrule
    ACDO \cite{AnimeColorDeOldify} &  &  0.8687 & 39.38 & 15.75 &0.7672&  4.540 &  0.8970 & 28.12 & 21.77 &0.9516 & 4.686\\ \midrule
    EBMC \cite{isola2017image} & \checkmark & 0.8542 & 38.77 & 15.21 &0.7592& 4.605 &  0.8859 & 19.48 & 20.80 &0.9474& 4.702\\ \midrule
    ScreenVAE \cite{xie2020manga} & \checkmark &  0.7328 & 98.52 & 9.12 &0.5373 & 4.160& - & - & - & -\\ \midrule
    Ours  & \checkmark &  \textbf{0.9419} & \textbf{13.37} & \textbf{25.88} &\textbf{0.9541}&\textbf{4.924}&  \textbf{0.9433} & \textbf{12.17} & \textbf{26.01} &\textbf{0.9579}&\textbf{5.011}\\
    \bottomrule[1.5pt]
  \end{tabular}
}
\vspace{-2mm}
\end{table*}

\paragraph{Training data.} 
The most direct application of sequence image colorization is in manga colorization. In this study, we compiled the largest manga colorization dataset to date, consisting of over 50,000 publicly available color manga chapter sequences sourced from various open online repositories, after filtering out black-and-white manga, resulting in more than 1.7 million images. For each manga frame, we randomly selected at least 20 additional frames from the corresponding manga chapter to construct a diverse reference image pool. Subsequently, we utilized the CLIP image encoder~\cite{radford2021learning} to identify and retrieve the 12 most relevant reference image patches. This systematic recording of selections facilitates subsequent training while minimizing redundant computations.

\paragraph{Evaluation Benchmark.} 
%
Previous reference-based colorization datasets, such as AnimateDiffusion\cite{cao2023animediffusion}, use a single colored image as a reference for each test image, typically featuring only one character. This limits the evaluation of multiple color ID preservation. In contrast, ColorFlow-Bench is the first benchmark organized by manga series, allowing models to reference multiple pages from the same manga for colorization. Moreover, this design effectively assesses a model's ability to preserve multiple color IDs, better aligning with industry needs. 
We establish the \BenchmarkName~ comprising 30 manga chapters that are not included in the training phase. Each chapter contains 40 reference images and 50 black-and-white manga pages, provided in two styles: screenstyle~\cite{xie2020manga} and grayscale image. With a total of 2,700 manga pages, \BenchmarkName~ stands as the largest manga colorization benchmark. 
It covers 22 Japanese manga and 8 Western comics, categorized into 11 webtoons and 19 multi-grid manga, ensuring a diverse and reliable dataset.
We evaluate the quality of the colorization and the fidelity of the colors to the original images using several metrics: CLIP Image Similarity (CLIP-IS)~\cite{radford2021learning}, Fréchet Inception Distance (FID)~\cite{fid}, Peak Signal-to-Noise Ratio (PSNR)~\cite{psnr}, Structural Similarity Index (SSIM)~\cite{ssim}, and Aesthetic Score (AS)~\cite{schuhmann2022laion}.
These metrics provide a thorough and holistic assessment of the colorization process, evaluating not only the aesthetic quality of the generated images but also their consistency with the original content.

\subsection{Implementation Details}

Our colorization model is based on Stable Diffusion v1.5~\cite{rombach2022high}. We trained our model, along with all ablation models, for 150,000 steps using 8 NVIDIA A100 GPUs, with a learning rate of 1e-5. Additionally, the Guided Super-Resolution Pipeline was trained for 30,000 iterations under the same hardware configuration and learning rate. For inference, all methods were tested on NVIDIA Tesla A100 GPUs, consistent with their open-source code.

\subsection{Baseline Models}
\label{sec:baseline}

To ensure a fair comparison, we select the most recent and competitive approaches in manga colorization. 
\textbf{Colorization without reference images} includes \textit{Manga Colorization V2 (MC-v2)}~\cite{mcv2}, which employs CycleGAN for automated manga colorization, and \textit{AnimeColorDeOldify (ACDO)}~\cite{acdo}, a DeOldify variant optimized for anime and manga.
\textbf{Colorization based on reference images} features \textit{Example Based Manga Colorization (EBMC)}~\cite{isola2017image}, which uses a cGAN to combine color features from reference images with grayscale content; \textit{ScreenVAE}~\cite{xie2020manga}, which utilizes variational autoencoders for colorization; and \textit{Style2Paints V4.5}~\cite{zhang2018two}, a software designed for coloring line art that matches color styles using reference images.





\subsection{Quantitative Comparisons}

\begin{figure}[t]
    \centering
    \includegraphics[width=0.48\textwidth]{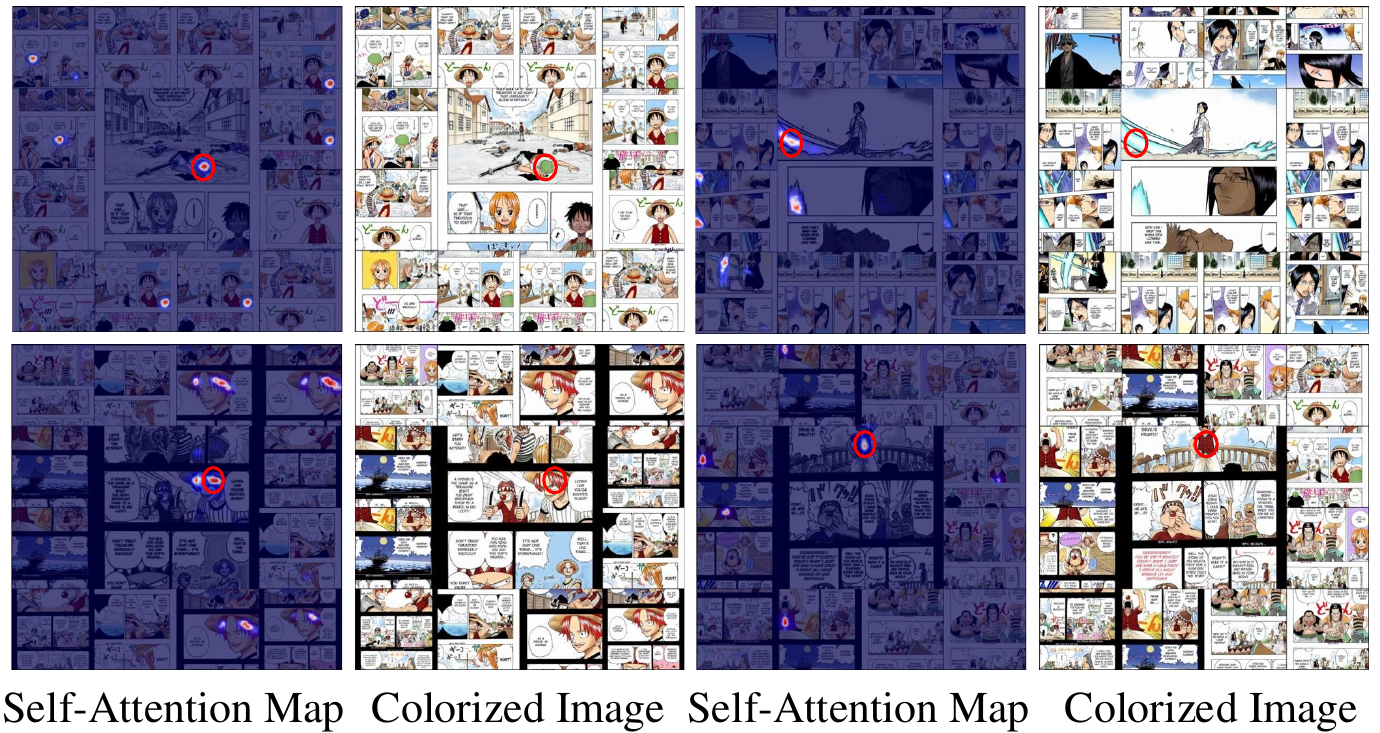}
    \vspace{-2mm}
    \caption{
    \textbf{Visualization of the heatmap} for the self-attention map of the selected colorization region (encircled in \red{red}).
    }
    \vspace{-2mm}
    \label{fig:attenmap}
\end{figure}

\begin{figure*}[t]
    \centering
    \vspace{-2mm}
    \includegraphics[width=0.99\textwidth]{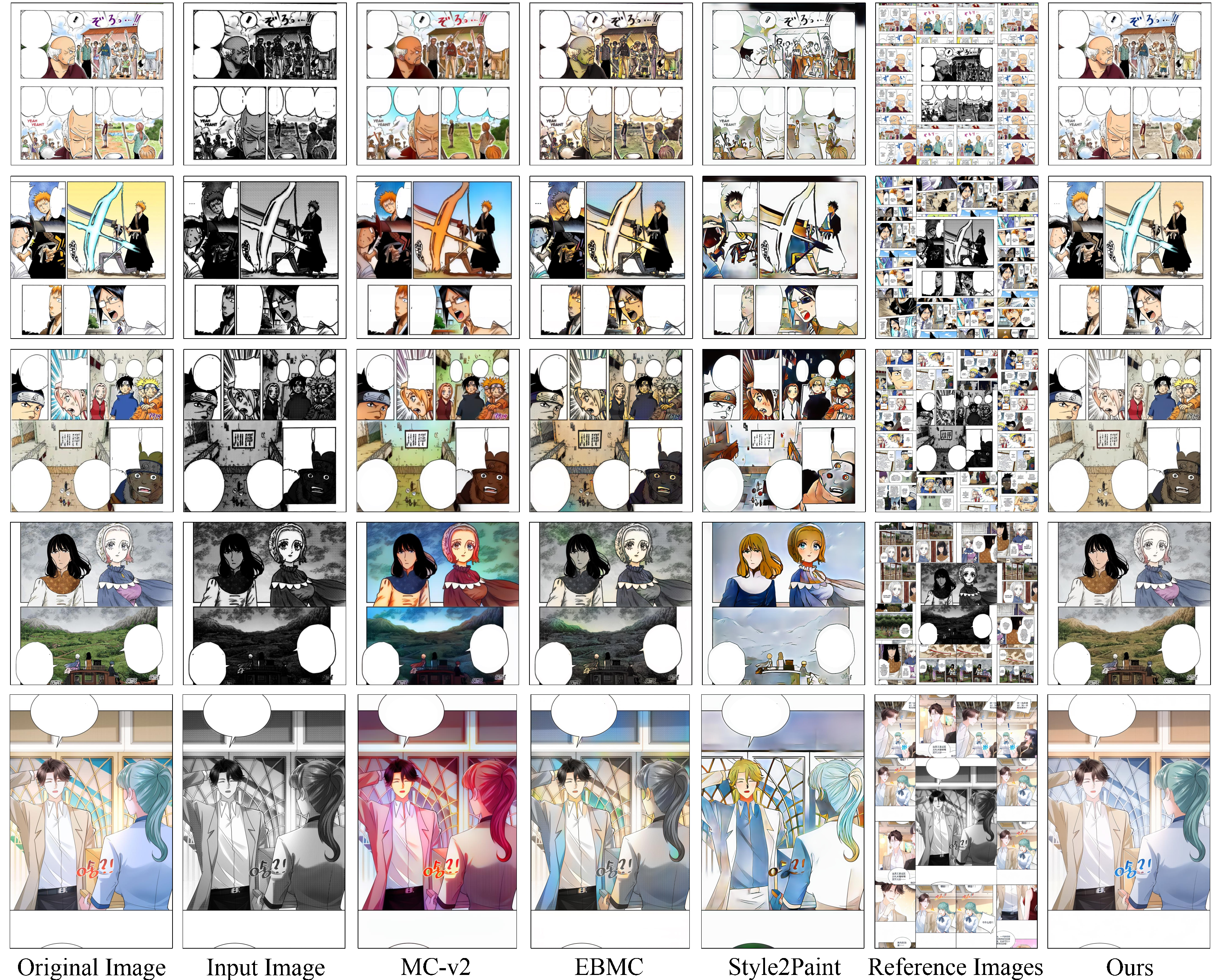}
    \vspace{-2mm}
    \caption{
    \textbf{Comparison of our method with SOTA approaches in the manga colorization.} Our method exhibits superior aesthetic quality, producing colors that more closely match the original image. {[Best viewed in color with zoom-in]}
    }
    \vspace{-1mm}
    \label{fig:exp1}
\end{figure*}


In Tab.~\ref{tab:exp1}, we present a comparison between \OurMethod\ and previous works using \BenchmarkName. Our results show that \OurMethod\ significantly outperforms other models across all metrics, including semantic alignment (CLIP-IS, FID), pixel alignment (PSNR, SSIM), and Aesthetic Score (AS), highlighting its superior image colorization accuracy. Although methods like EBMC \cite{isola2017image} and ScreenVAE \cite{xie2020manga} are capable of reference-based colorization, they fall short due to weak in-context learning and an inability to maintain consistent sequential colorization. In contrast, \OurMethod\ excels by utilizing diffusion models to effectively preserve color identity, as demonstrated by the self-attention map in Fig. \ref{fig:attenmap}.
\subsection{Qualitative Comparisons}
\begin{figure*}[t]
    \centering
    \includegraphics[width=1.\textwidth]{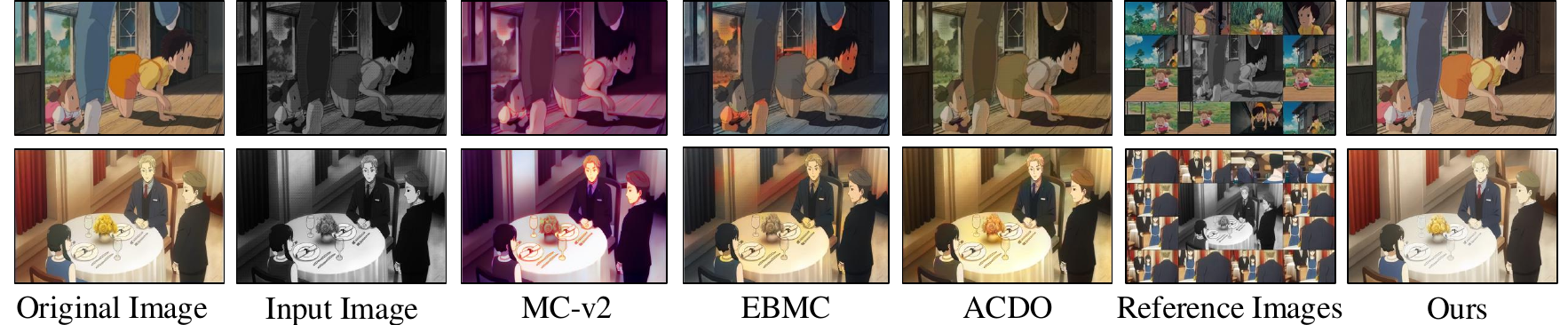}
    \vspace{-0.2cm}
    \caption{
    \textbf{Comparison of ColorFlow with other approaches in the animation storyboard colorization.} Our method exhibits superior aesthetic quality, producing colors that more closely match the original image. {[Best viewed in color with zoom-in]}
    }
    \vspace{-0.2cm}
    \label{fig:exp2}
\end{figure*}
\begin{figure}[t]
    \centering
    \includegraphics[width=0.48\textwidth]{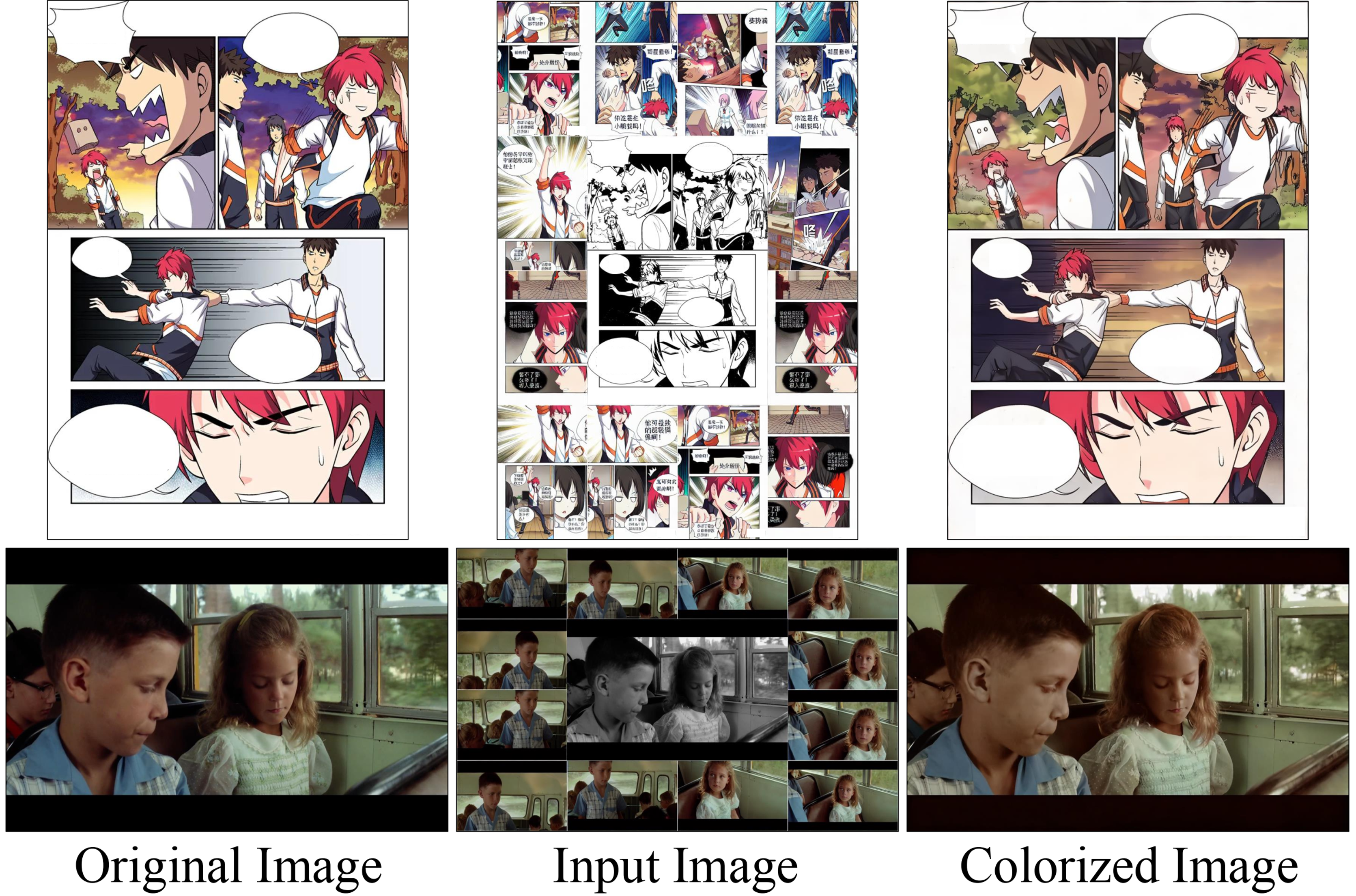}
    \vspace{-0.5cm}
    \caption{
       Colorization results for line art and natural scenario.
    }
    \label{fig:nature}
\end{figure}
To showcase \OurMethod's generalization ability, we present qualitative results in four scenarios: manga colorization (Fig.\ref{fig:exp1}), cartoon colorization (Fig.~\ref{fig:exp2}), line art colorization  (Fig.~\ref{fig:nature}), and natural scenes colorization (Fig.~\ref{fig:nature}).

Fig.~\ref{fig:exp1} compares the colorization results of \OurMethod\ with previous methods. The no-reference model MC-v2\cite{mcv2} lacks contextual awareness, leading to random colorization. EBMC and Style2Paints\cite{zhang2018two} use reference images but suffer from information loss, resulting in imprecise colorization. In contrast, \OurMethod~integrates reference images effectively using image stitching, leveraging self-attention layers in diffusion models to maintain color consistency across manga frames.

Fig.~\ref{fig:exp2} and Fig.~\ref{fig:nature} demonstrate the exceptional performance of \OurMethod\ across a diverse range of scenarios, including cartoon, line art, and natural images.  These results emphasize the robustness and adaptability of our approach, demonstrating its strong generalization capabilities to effectively handle varying styles and content types. 

\subsection{Abaltion Study}

\vspace{-0.1cm}

\paragraph{Pipeline components. } In Tab.~\ref{tab:SRP-RP}, we compare the impact of the Retrieval-Augmented Pipeline and the Guided Super-Resolution Pipeline during training and inference. 
The results indicate that using the Retrieval-Augmented Pipeline for both training and inference, as well as the Guided Super-Resolution Pipeline during training, is crucial for the performance of \OurMethod.

\begin{table}[htbp]
    \centering
  \caption{
  Ablation Study on the Influence of Retrieval-Augmented Pipeline (RAP) and Guided Super-Resolution Pipeline (GSRP). 
  }
  \vspace{-1mm}
  \label{tab:SRP-RP}
\resizebox{0.46\textwidth}{!}{
      \begin{tabular}{c | c c| c c c c c}
    \toprule[1.5pt]
     Training & \multicolumn{2}{c|}{Inference}  & \multirow{2}{*}{CLIP-IS$\uparrow$} & \multirow{2}{*}{FID$\downarrow$} & \multirow{2}{*}{PSNR$\uparrow$} & \multirow{2}{*}{SSIM$\uparrow$} &\multirow{2}{*}{AS$\uparrow$}\\ 
     RAP & RAP & GSRP &   &   &  & \\ \midrule
     & \checkmark & \checkmark &  0.9326 & 15.98 & 24.48 & 0.9448 & 4.921\\ \midrule
    &  & \checkmark &  0.9233 & 18.32 & 24.16 &0.9410 & 4.907\\ \midrule
    \checkmark &  & \checkmark &  0.9266 & 17.07 & 24.64 &0.9464 & 4.914\\ \midrule
    \checkmark & \checkmark &  &  0.9322 & 17.85 & 20.12 & 0.8077 & 4.898\\ \midrule
    \checkmark & \checkmark & \checkmark &  \textbf{0.9419} & \textbf{13.37} & \textbf{25.88} &\textbf{0.9541}&\textbf{4.924}\\ 
    \bottomrule[1.5pt]
  \end{tabular}
  \vspace{-5mm}
}
\end{table}


\paragraph{Inference resolution.} In Tab.~\ref{tab:resolution}, we conduct an ablation study on three different inference resolutions. Despite being trained only on a resolution of $512\times800$, the results demonstrate that \OurMethod~ has the ability to generalize across different resolutions.

\begin{table}[htbp]
    \centering
  \caption{
  Ablation Study of Inference Resolution.
  }
  \vspace{-1mm}
  \label{tab:resolution}
\resizebox{0.48\textwidth}{!}{
      \begin{tabular}{c | c c c c c}
    \toprule[1.5pt]
     Width $\times$ Height (Pixel) &  CLIP-IS$\uparrow$ & FID$\downarrow$  & PSNR$\uparrow$ &SSIM$\uparrow$ &AS$\uparrow$\\ \midrule
    512 $\times$ 800 &  0.9372  &  14.91 &  23.51 & 0.9414 & 4.868\\ \midrule
    1024 $\times$ 1600  &  \textbf{0.9419} & \textbf{13.37} & 25.88 &0.9541 & 4.924\\ \midrule
    1280 $\times$ 2000  &  0.9398 & 13.42 &  \textbf{26.02} & \textbf{0.9580}& \textbf{4.929}\\ 
    \bottomrule[1.5pt]
  \end{tabular}
  \vspace{-5mm}
}
\end{table}

\vspace{-6mm}

\paragraph{LoRA rank.} To demonstrate the necessity of partially retaining pretrained diffusion weights, we ablate on the rank of LoRA on the base diffusion model, where larger LoRA rank indicates bigger change on pretrained diffusion model weight. Tab.~\ref{tab:lora} shows that too big or too small LoRA rank all lead to performance decay, validating the choice of 64 for the optimal LoRA rank.

\begin{table}[htbp]
    \centering
    \vspace{1mm}
  \caption{
  Ablation Study of LoRA Rank.
  }
  \vspace{-1mm}
  \label{tab:lora}
\resizebox{0.35\textwidth}{!}{
      \begin{tabular}{c | c c c c c}
    \toprule[1.5pt]
     Rank &  CLIP-IS$\uparrow$ & FID$\downarrow$  & PSNR$\uparrow$ &SSIM$\uparrow$ &AS$\uparrow$\\ \midrule
    32 & 0.940 & 13.46 & 25.46 & 0.9521 & 4.920\\ \midrule
    64  &  \textbf{0.9419} & \textbf{13.37} & \textbf{25.88} &\textbf{0.9541}&{4.924}\\ \midrule
    128  &  0.9376 & 14.31 & 24.79 & 0.9461 & \textbf{4.930}\\ \midrule
    192  &  0.9370 & 14.46 & 24.59 & 0.9440 & 4.914\\
    \bottomrule[1.5pt]
  \end{tabular}
}
\end{table}

\vspace{-3mm}

\paragraph{Sampling timesteps.} In Tab.~\ref{tab:timesteps}, we ablate on the design of timestep shift sampling. Since colorization is primarily performed at a higher timestep, we strengthen the sampling at higher timestep by a factor of $\mu$. The results validate the effectiveness of adding timesteps sampling and using a factor of $\mu = 1.5$.

\begin{table}[htbp]
    \centering
  \caption{
  Ablation Study of Timesteps Sampling.
  }
  \vspace{-1mm}
  \label{tab:timesteps}
\resizebox{0.34\textwidth}{!}{
      \begin{tabular}{c | c c c c c}
    \toprule[1.5pt]
     $\mu$ &  CLIP-IS$\uparrow$ & FID$\downarrow$  & PSNR$\uparrow$ &SSIM$\uparrow$ &AS$\uparrow$\\ \midrule
    0 &  0.9351  &  14.18 & 25.12  & 0.9501 & \textbf{4.927} \\ \midrule
    1.5  &  \textbf{0.9419} & \textbf{13.37} & \textbf{25.88} &\textbf{0.9541} & 4.924\\ \midrule
    3  &  0.9395 & 13.51 & 25.42 & 0.9509 & 4.917\\ 
    \bottomrule[1.5pt]
  \end{tabular}
}
\end{table}

\vspace{-2.8mm}

\subsection{User Study}
To perform a comprehensive comparison, we conducted a user study evaluating three key aspects: aesthetic quality, similarity to the original image, and consistency of color IDs in image sequences.
In each trial, participants ranked their preferences among five sample groups. We assigned scores based on these rankings, with the first place receiving 5 points and the fifth place receiving 1 point. We then calculated the average score for each evaluation criterion. As detailed in Tab.~\ref{tab:user}, we gathered over 4,000 valid rankings. The results demonstrate that our colorization method is the preferred choice across all evaluation criteria.
\begin{table}[htbp]
    \centering
  \caption{
  Results of the User Study. The table presents the average \textbf{Score} for different models based on aesthetic quality, similarity to the original image, and consistency in sequences 
  }
  \vspace{-2mm}
  \label{tab:user}
\resizebox{0.48\textwidth}{!}{
      \begin{tabular}{c | c c c c c}
    \toprule[1.5pt]
      &  Ours  & EBMC &MC-v2 &ACDO & ScreenVAE\\ \midrule
    Aesthetic Quality $\uparrow$ &  \textbf{4.577} &  3.141 & 2.891  & 2.844 & 1.547 \\ \midrule
    Similarity to Original  $\uparrow$  &  \textbf{4.673} & 3.316 & 2.984 &2.642 & 1.385\\ \midrule
    Consistency in Sequences $\uparrow$  &  \textbf{4.538} & 3.399 & 3.215 & 2.540 & 1.308\\ 
    \bottomrule[1.5pt]
  \end{tabular}
}
\end{table}
\vspace{-3mm}

\section{Conclusion}
\label{sec:conclusion}

In conclusion, this paper proposes \OurMethod\ for a novel task, Reference-based Image Sequence Colorization. The proposed method consists of a three-stage framework: the Retrieval-Augmented Pipeline, the In-context Colorization Pipeline, and the Super-Resolution Pipeline. Extensive quantitative and qualitative evaluation results on our proposed benchmark, \BenchmarkName, show the superior performance of \OurMethod. More discussion on limitations and future work will be discussed in supplementary files.

{
    \small
    \bibliographystyle{ieeenat_fullname}
    \bibliography{main}
}

\clearpage
\setcounter{page}{1}
\maketitlesupplementary



\noindent This supplementary material provides further insights and additional results. It elaborates on ColorFlow's performance across various artistic contexts, discussing retrieval effectiveness, the model's limitations, and the ethical considerations associated with generating synthetic content.



\vspace{-0.2cm}

\section{Additional Results}
\label{sec:additional_results}

\vspace{-0.1cm}

\begin{figure*}[t]
    \centering
    \vspace{-2mm}
    \includegraphics[width=0.95\textwidth]{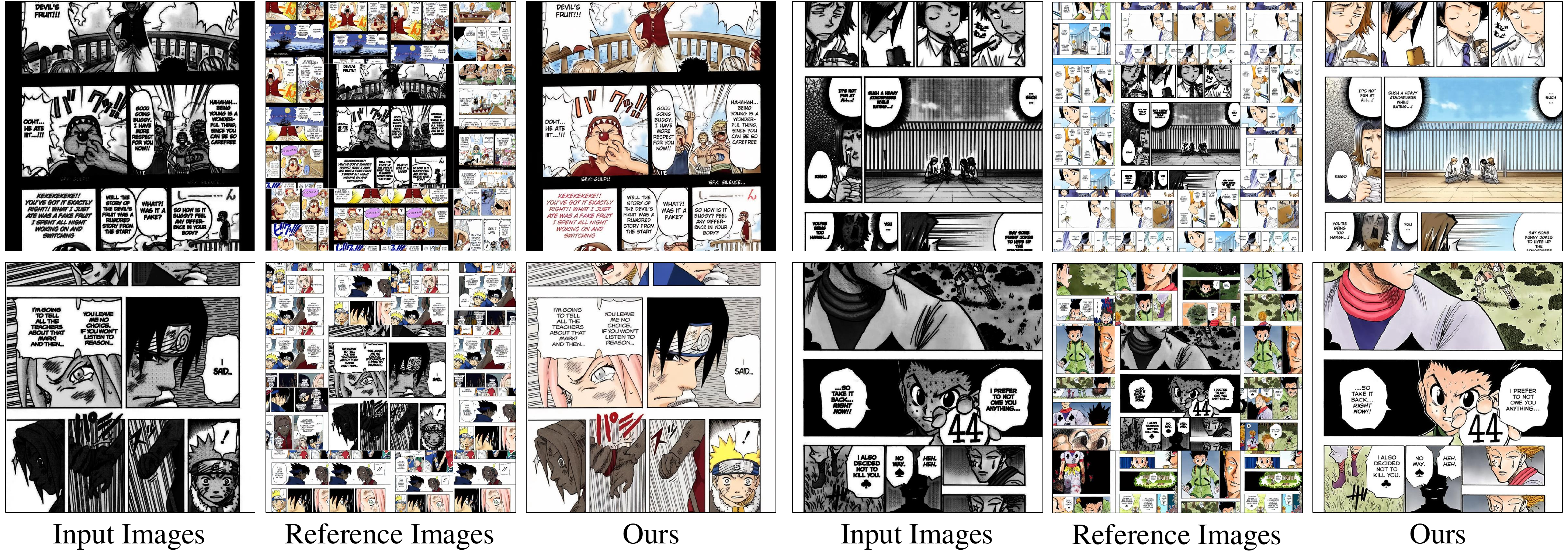}
    \vspace{-2mm}
    \caption{
    \textbf{Colorization results of black and white manga using ColorFlow.} {[Best viewed in color with zoom-in]}
    }
    \vspace{-1mm}
    \label{fig:sup_manga}
\end{figure*}

\begin{figure*}[t]
    \centering
    \vspace{-2mm}
    \includegraphics[width=0.95\textwidth]{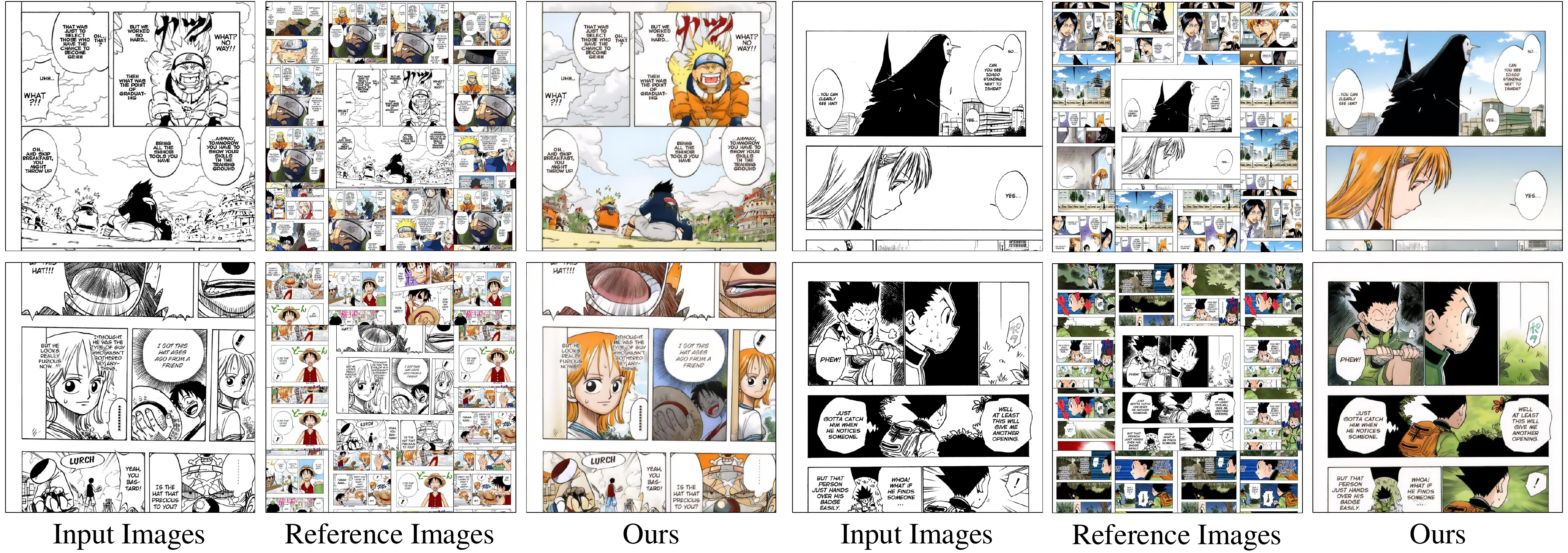}
    \vspace{-2mm}
    \caption{
    \textbf{Colorization results of line art using ColorFlow.} {[Best viewed in color with zoom-in]}
    }
    \vspace{-1mm}
    \label{fig:sup_line}
\end{figure*}

\begin{figure*}[t]
    \centering
    \vspace{-2mm}
    \includegraphics[width=0.95\textwidth]{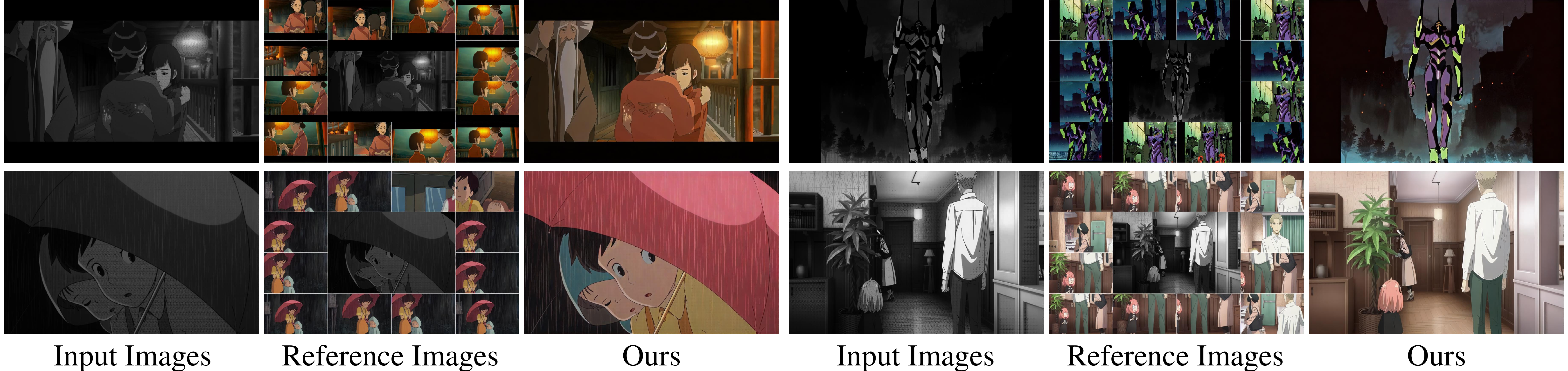}
    \vspace{-2mm}
    \caption{
    \textbf{Colorization results of animation storyboard using ColorFlow.} {[Best viewed in color with zoom-in]}
    }
    \vspace{-1mm}
    \label{fig:sup_cartoon}
\end{figure*}

\begin{figure*}[htp!]
    \centering
    \vspace{-2mm}
    \includegraphics[width=0.95\textwidth]{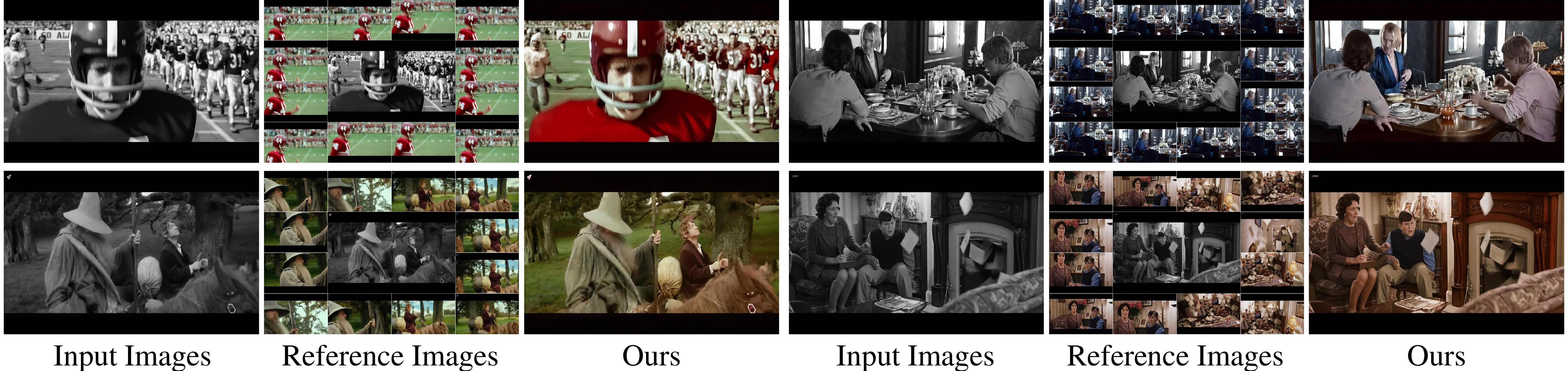}
    \vspace{-2mm}
    \caption{
    \textbf{Colorization results of natural scenario using ColorFlow.} {[Best viewed in color with zoom-in]}
    }
    \vspace{-1mm}
    \label{fig:sup_real}
\end{figure*}

To highlight the robustness and versatility of ColorFlow, we present a series of visual results across diverse artistic contexts. Fig.\ref{fig:sup_manga}, \ref{fig:sup_line}, \ref{fig:sup_cartoon}, and \ref{fig:sup_real} demonstrate ColorFlow's performance on black-and-white manga, line art, animation storyboards, and grayscale natural scenes, respectively. 
These examples collectively emphasize ColorFlow's adaptability across various contexts, establishing it as a valuable tool for artists and content creators aiming to enhance their work through automated colorization.

\vspace{-0.2cm}

\section{Effectiveness of CLIP Retrieval}
\begin{figure}[t]
    \centering
    \vspace{-0.2cm}
    \includegraphics[width=0.4\textwidth]{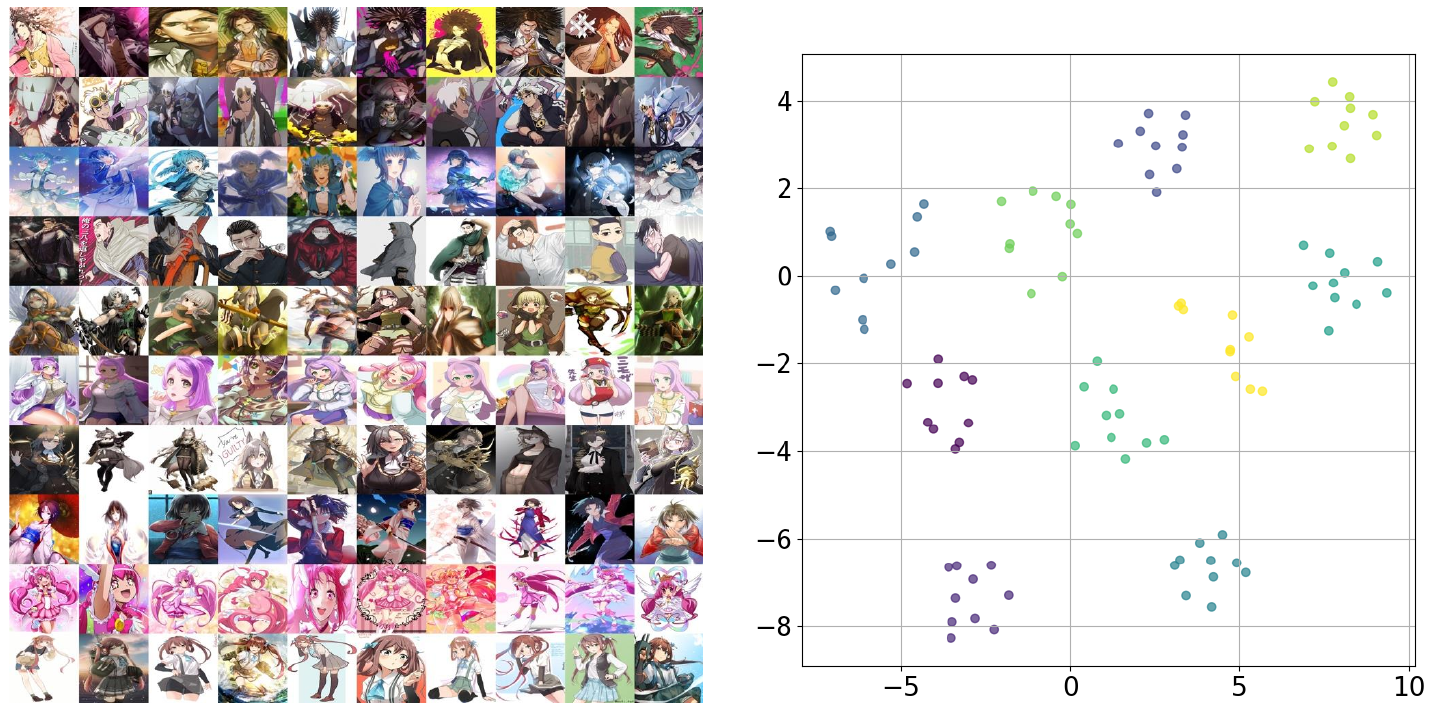}
    \vspace{-0.2cm}
    \caption{
       Visualization of 10 characters, each represented by 10 images (left). The corresponding t-SNE plot illustrates the CLIP image embeddings (right).
    }
    \vspace{-0.6cm}
    \label{fig:tsne}
\end{figure}

We collected 10 anime characters from the internet, each with 10 images. A t-SNE visualization of the CLIP embeddings \footnote{Using the CLIP-ViT-H-14-laion2B-s32B-b79K model} shows that images of the same character cluster together (see Figure \ref{fig:tsne}). In top-1 retrieval tests, we achieved a 100\% success rate with color images and 97\% with grayscale. These results demonstrate that the CLIP image encoder effectively retrieves the corresponding characters,  highlighting the robustness of the RAP.

\section{Limitations and Future Work}
\label{sec:limitations}

\vspace{-0.1cm}

Despite the significant advancements achieved by ColorFlow in reference-based image sequence colorization, several limitations require careful consideration.

First, ColorFlow's performance heavily relies on the quality of images in reference pool. If the artistic style of reference images is highly abstract or significantly different from the target style, the colorization accuracy may suffer.

Additionally, ColorFlow's ability to generate images and preserve color identity is limited by its base model, Stable Diffusion 1.5~\cite{rombach2022high}. Although this model is effective, there is room for improvement by using more advanced architectures like Flux.1 or SD3~\cite{esser2024scaling}. In future work, we plan to train ColorFlow on these next-generation models, which could enhance color fidelity and overall image quality.

We also intend to explore using ColorFlow for long video colorization. This would expand its use in multimedia production, allowing for consistent colorization across extended video frames.
\vspace{-0.2cm}
\section{Ethical Considerations}
\label{sec:ethical_considerations}

\vspace{-0.1cm}

While our research primarily emphasizes the technical advancements in image colorization, it is acknowledged there are ethical implications associated with the generation of synthetic content. We address the limitation in this section.
%

Our model is trained on data sourced from the internet, which may inadvertently reflect and amplify existing biases present in the training data. This concern can be salient, as artificial intelligence systems trained on biased datasets can perpetuate stereotypes and exacerbate inequalities, disproportionately affecting various demographic groups. To address this issue, we tried to ensure diversity in the training data to cover a wide range of styles, demographics, and cultural contexts. Moreover, we will monitor and evaluate the model for biased behavior, and fine-tune it using balanced datasets. Moreover, it is possible that the colorization method could be used maliciously, such as altering historical artifacts or creating misleading media. To address this issue, we will include watermarking or traceable signatures in outputs to indicate AI-generated content.
We will also publish ethical usage guidelines for the model and monitor public applications of the model to identify and address misuse. By addressing these issues, our work can maintain ethical integrity while maximizing its positive impact.


\vspace{-0.2cm}




\end{document}